\documentclass[runningheads]{llncs}
\usepackage{amssymb}
\usepackage{tablefootnote}
\usepackage{graphicx}
\usepackage{todonotes}
\usepackage{booktabs}
\usepackage{multirow}
\usepackage{hyperref}
\usepackage{subcaption}
\usepackage{float}
\usepackage[export]{adjustbox}
\usepackage{tablefootnote}
\usepackage[utf8]{inputenc}
\usepackage[para,online,flushleft]{threeparttable}
\newcommand{\block}[1]{%
  \raisebox{\dimexpr(\fontcharht\font`X-1em)/2}{\rule{1em}{#1\dimexpr1em/8}}%
}
\DeclareUnicodeCharacter{2581}{\block{1}}
\usepackage{pifont}
\newcommand{\cmark}{\ding{51}}%
\definecolor{green}{HTML}{32CD32}
\definecolor{red}{HTML}{FF5733}

\begin{document}

\title{Improving Automatic Text Recognition with Language Models in the PyLaia Open-Source Library}
\titlerunning{Improving the PyLaia open-source library}

\author{
    Solène Tarride\orcidID{0000-0001-6174-9865} 
    \and Yoann Schneider 
    \and Marie Generali-Lince
    \and Mélodie Boillet\orcidID{0000-0002-0618-7852}
    \and Bastien Abadie 
    \and Christopher Kermorvant\orcidID{0000-0002-7508-4080} 
}
\authorrunning{Tarride et al.}
%
\institute{TEKLIA, Paris, France}

\maketitle              

\begin{abstract}
PyLaia is one of the most popular open-source software for Automatic Text Recognition (ATR), delivering strong performance in terms of speed and accuracy. 
In this paper, we outline our recent contributions to the PyLaia library, focusing on the incorporation of reliable confidence scores and the integration of statistical language modeling during decoding. Our implementation provides an easy way to combine PyLaia with $n$-grams language models at different levels. One of the highlights of this work is that language models are completely auto-tuned: they can be built and used easily without any expert knowledge, and without requiring any additional data. 
To demonstrate the significance of our contribution, we evaluate PyLaia's performance on twelve datasets, both with and without language modelling. The results show that decoding with small language models improves the Word Error Rate by 13\% and the Character Error Rate by 12\% in average. Additionally, we conduct an analysis of confidence scores and highlight the importance of calibration techniques.
Our implementation is publicly available in the official PyLaia repository (\url{https://gitlab.teklia.com/atr/pylaia}), and twelve open-source models are released on Hugging Face\footnote{\url{https://huggingface.co/collections/Teklia/pylaia-65f16e9ae0aa03690e9e9f80}}.


\end{abstract}




\section{Introduction}

The scope of Automatic Text Recognition (ATR) has expanded significantly in recent years as deep learning models have become more robust and efficient. These advances have contributed to the development of easy-to-use web platforms, such as Transkribus\footnote{\url{https://readcoop.eu/transkribus/}}, Transkriptorium\footnote{\url{http://www.transkriptorium.com/}}, eScriptorium\footnote{\url{https://escriptorium.fr/}} or Arkindex\footnote{\url{https://doc.arkindex.org/}}, allowing non machine-learning experts to train deep learning models and process multiple documents with a single click. These platforms often rely on open-source ATR software, such as PyLaia\footnote{\url{https://gitlab.teklia.com/atr/pylaia}} which is one of the engines available in Transkribus and Arkindex. 

Introduced in 2017, PyLaia \cite{laia} is an open-source library for automatic text recognition that allows the training of convolutional recurrent neural networks with the CTC loss function. As an optical model, PyLaia makes decisions based on the shape of letters in images. 
On the other hand, n-gram language models, trained on a text corpus, estimate a probability distribution for the next token (character, word, or subword) given a context of size $n$. Combining these technologies allows decisions to be made based on both visual and linguistic context, and could improve PyLaia's performance, as suggested by Maarand \textit{et al.} \cite{HuMu-opensourceHTR}.

Our implementation facilitates the integration of PyLaia with n-gram language models at the character, subword, or word level, supporting models built with the SRILM\footnote{\url{http://www.speech.sri.com/projects/srilm/}} or KenLM\footnote{\url{https://github.com/kpu/kenlm}} toolkits. This combination significantly improves handwriting recognition results. In particular, the language models are automatically tuned for ease of use without the need for expert knowledge or additional data. We also introduce calibrated confidence scores using temperature scaling to increase the reliability of PyLaia's output.
To summarize, our contributions to the PyLaia library include:
\begin{itemize}
    \item Confidence score calibration with temperature scaling ;
    \item Language model support, either at character, subword and word levels ;
    \item A comprehensive documentation on how to use and contribute to PyLaia\footnote{\url{https://atr.pages.teklia.com/pylaia/}} ;
    \item The release of 12 open-source models for ATR, covering 6 languages\footnote{\url{https://huggingface.co/Teklia}} ;
    \item The publication of 12 existing datasets on Hugging Face, with an easy way to convert them in PyLaia format.
\end{itemize}

This paper is organized as follows. The next Section \ref{sec:related_works} gives an overview of research related to language models for ATR. In Section \ref{sec:pylaia} we introduce the PyLaia library and present our contributions in Section \ref{sec:contributions}. In Section \ref{sec:experimental-setup}, we describe the datasets and parameters used in our experiments. Finally, the results of our contributions are presented and discussed in Section \ref{sec:results}.  

\section{Related works}
\label{sec:related_works}
In this section, we present research that addresses the significance of language models in improving Automatic Text Recognition (ATR).

\subsection{Open-source Automatic Text Recognition libraries}
In 2022, Maarand \textit{et al.} conducted a survey of ATR libraries used in recent scientific peer-reviewed articles published in the document processing community \cite{HuMu-opensourceHTR}. Their study identified four open-source libraries for ATR at text-line level that were still actively maintained at the time, based on contributors and recent commits.

\begin{itemize}
    \item Kaldi\footnote{\url{https://github.com/kaldi-asr/kaldi}} \cite{kaldi} is a library developed for speech recognition and adapted to ATR. To this day, Kaldi is still actively maintained.
    \item Kraken\footnote{\url{https://github.com/mittagessen/kraken}} \cite{kraken} is a ready-to-use OCR system specifically optimized for historical and non-Latin script material. Although it was designed for printed text, it can also be trained to recognize cursive and handwritten scripts. To this day, Kraken is still actively maintained.
    \item PyLaia\footnote{\url{https://gitlab.teklia.com/atr/pylaia}} \cite{laia} is a deep learning toolkit for handwritten document analysis based on PyTorch. It is one of the ATR engines available in Transkribus. To this day, PyLaia is still actively maintained. 
    \item HTR-Flor++\footnote{\url{https://github.com/arthurflor23/handwritten-text-recognition}} \cite{HTR-Flor++} is a framework for ATR that implements different state-of-the-art architectures, based on TensorFlow. Unfortunately, active development for HTR-Flor++ stopped at the end of 2022.
\end{itemize}

Since this study was published, three other open-source models based on Transformers were made available for Automatic Text Recognition. 
\begin{itemize}
    \item DAN \cite{dan} is composed of a fully convolutional encoder and a text Transformer decoder. It operates at any level, e.g. text-lines, paragraphs, or pages. The development is not active, as the last commit was in April 2023. 
    \item TrOCR \cite{trOCR} consists of an image Transformer encoder and an autoregressive text Transformer decoder. It operates on text-lines. TrOCR is fully integrated in Hugging Face.
    \item MMOCR \cite{mmocr2021} is an open-source OCR toolbox based on PyTorch designed for printed text detection, recognition, and other downstream tasks including key information extraction. It includes several models trained for each task, including Transformers.
    
\end{itemize}

Transformer-based models often yield higher performance, but are notoriously harder to train since they require a lot of data and computing power. Additionally, their inference time tends to be higher compared to other architectures.


\subsection{Improving Automatic Text Recognition models with language models}
Language models play a crucial role in improving the performance of Automatic Text Recognition (ATR) systems, especially in the case of noisy documents or ambiguous writing styles.
Language models can help by examining the previous words, subwords, or characters, enabling ATR systems to make more contextually informed decisions.

In recent model architectures, language modeling is implicitly learned by the decoder. Traditional Recurrent Neural Networks (RNNs) manage sequential data by maintaining hidden states to preserve the context of previously predicted tokens. However, they face challenges such as vanishing gradient problems and a limited ability to capture rich context. To address these issues, advanced architectures such as Long Short-Term Memory (LSTM) and Gated Recurrent Units (GRUs) have been introduced to improve the capture of long-range dependencies, proving particularly effective in tasks such as automatic text recognition (ATR) \cite{laia}.
More recently, the integration of attention mechanisms by Transformer models has sparked a significant shift in natural language processing (NLP). Transformers excel at modeling long sequences and have become the cornerstone of high-performance models in several applications, including ATR \cite{nougat,dan,trOCR}.

Because ATR often faces challenges with distorted or noisy documents, researchers have also relied on explicit statistical language models to improve the performance of CTC-based neural networks \cite{kumar2017lattice,laia,zhang2020scut}.
The most common form of explicit models used in this context are n-gram models, which are probabilistic models that capture the statistical relationships between sequences of tokens in natural language. They are based on the assumption that the probability of the next token in a sequence depends only on a fixed-sized window of previous tokens. Smoothing strategies are often applied to avoid assigning zero probability to tokens that have not been encountered before. These statistical models have been successful in a variety of language-related tasks. 
Explicit language models take into account the probability of text sequences and can be used as a prior in the decoding process to improve recognition. This context awareness allows ATR systems to select more contextually appropriate interpretations, resulting in improved recognition. N-gram models are mainly trained at character \cite{rethinking-line-atr,zhang2020scut,n-grams-in-atr} and word \cite{laia,n-grams-in-atr} levels.

\section{PyLaia}
\label{sec:pylaia}
PyLaia is an open-source tool written in Python, and designed for Automatic Text Recognition (ATR) at text-line level. The PyLaia library allows users to train and predict any CNN-RNN architecture trained with the CTC loss.




\subsection{Commands description}
Three commands are currently available in PyLaia: model creation, model training, and prediction.

\subsubsection{Model creation}

The \texttt{pylaia-htr-create-model} command creates a model and saves it as a pickle file. 
Models created with PyLaia are fully configurable by the user via command line arguments or a YAML configuration file. The general architecture of PyLaia is composed of convolutional blocks followed by a set of bidirectional recurrent layers and a final linear layer. Different parameters can be set by the user: number of convolutional blocks, number of recurrent layers, whether to use batch normalization, pooling layers, or activation function.

\subsubsection{Training}
The \texttt{pylaia-htr-train-ctc} command trains a PyLaia model and saves the best and last weights at each epoch. Many parameters can be set by the users via command line arguments or a YAML configuration file: the seed, batch size, early stopping strategy, checkpointing mode, data augmentation, optimizer and scheduler, etc. The full list is described in the documentation.

\subsubsection{Prediction}
The \texttt{pylaia-htr-decode-ctc} command applies a trained PyLaia model on a set of images, with or without language models. Many parameters can be set by the user, including text formatting and the visualization of CTC alignments. We have also added parameters to display confidence scores, calibrate confidence scores, adjust language model parameters and lexicon specifications.

\subsection{Code quality}
After the migration to GitLab, we added some DevOps tools to PyLaia in order to maintain a high level of code quality.

\begin{itemize}
    \item \textbf{Code linting and formatting}. The codebase already followed Black\footnote{\url{https://black.readthedocs.io/en/stable/}}'s formatting style and Isort\footnote{\url{https://pycqa.github.io/isort/}}'s guidelines for import sorting. We simply migrated this setup to use Ruff\footnote{\url{https://docs.astral.sh/ruff/}}, which includes both rule-sets and more. Written in Rust, it is much faster than its Python counterparts while being easier to configure.
    \item \textbf{Unit tests}. PyLaia already included many unit tests to detect regressions and bugs. They are launched via Pytest\footnote{\url{https://docs.pytest.org/en/7.4.x/}} but we moved to Tox\footnote{\url{https://tox.wiki/en/latest/index.html}} which is better suited when managing multiple environments.
    \item \textbf{Documentation}. We have set up a public documentation\footnote{\url{https://atr.pages.teklia.com/pylaia/}}. This documentation is hosted on GitLab pages and generated using the MkDocs\footnote{\url{https://mkdocs.readthedocs.io/en/stable/}} ecosystem. Users can find information on the commands, the code reference, as well as changes provided by each new release.
    \item \textbf{Releases}. The PyLaia library can be installed using pip\footnote{\url{https://pypi.org/project/pylaia/}}. Releases are deployed whenever a tag is pushed on the repository. A GitLab release\footnote{\url{https://gitlab.teklia.com/atr/pylaia/-/releases}} is also created automatically, which includes all the new changes included in the release. A more digest version of the changes\footnote{\url{https://atr.pages.teklia.com/pylaia/releases/}} is also published on the public documentation.
    \item \textbf{Docker}. If developers want to use the library in Docker containers\footnote{\url{https://www.docker.com/}}, a Docker image is available for each tagged version. 
    \item \textbf{Dependencies}. Merge requests are automatically opened to bump dependencies.
    \item \textbf{CI/CD}. All these operations are launched at each commit. 
\end{itemize}

\section{Our contributions to the PyLaia library}
\label{sec:contributions}

In this section, we outline our major contributions to the PyLaia library, which include the integration of language models and the inclusion of calibrated confidence scores.
We also list the resources that have been released for PyLaia.

\subsection{Confidence scores}

Confidence scores are essential to assess the quality of any prediction in an unsupervised way. However, PyLaia did not output confidence scores until 2022. Therefore, our main focus was to implement confidence scores during decoding. 

\subsubsection{Confidence estimation methods}

We explore multiple ways of computing confidence scores for ATR.


\begin{itemize}
    \item \textbf{Softmax mean (before CTC)}: This method involves calculating the average of the highest softmax probability across all frames, including frames associated with blank and duplicate tokens.
    \item \textbf{Softmax mean (after CTC)}: This one is very similar to the previous one. The average of the highest softmax probability over all frames is computed. However, frames associated with blank and duplicate tokens are excluded from the calculation.
    \item \textbf{Softmax difference}: In this approach, we quantify the difference between the maximum probability and the second maximum probability for each frame. A high value indicates a more distinct separation between the most likely token and potential challengers.
    \item \textbf{Softmax entropy}: This method computes the average entropy over all frames, providing insight into the overall uncertainty or randomness in the predicted probabilities.
    \item \textbf{Monte Carlo Dropout} \cite{monte-carlo-dropout}: This last approach consists in enabling dropout during inference, in order to obtain $N$ predictions for each sample. The variance of the different logits for a single sample is then used to estimate a confidence score. 
\end{itemize}

Our statement is that an informative confidence score should provide an estimate of the recognition quality. As a result, confidence scores and the Character Recognition Rate should be correlated. In Section \ref{sec:res_conf}, we measure this correlation for each computation method and discuss the results. 

\subsubsection{Calibrating confidence scores}
We observed that PyLaia tends to be overconfident, with most confidence scores falling in the [0.97-0.99] range. In fact, neural networks are known to be poorly calibrated \cite{tempscaling}. To solve this problem, Guo \textit{et al.} recommend using temperature scaling. As a result, we explore how this technique can help calibrate PyLaia's confidence scores.

Temperature scaling is an extension of the Platt scaling calibration method adapted for multiclass classification. The problem of low standard deviation of the confidence score distribution can be attributed to the saturation behavior of the softmax function. A single parameter $T$, named temperature, can be used to increase the output entropy in order to soften the softmax. 

The idea is to scale the logits, a vector $x$ of size $n$, to get back into the unsaturated zone of the softmax by replacing the softmax function by
\begin{equation}
\sigma(x_i, T) = \frac{e^{x_i / T}}{\sum_{j=1}^{n} {e^{x_j / T}}}
\end{equation}

Note that, since the temperature is constant over the logits, it scales the output of the softmax function in a way that does not alter the rank nor the distribution of the logits. The only effect of this approach is to improve the interpretability of confidence scores for humans.

In Section \ref{sec:res_calibration}, we compare different values for the temperature and discuss its impact on confidence scores. 

\subsection{Explicit language modeling}
In their recent survey, Maarand \textit{et al.} \cite{HuMu-opensourceHTR} suggested that although PyLaia achieves high efficiency, the addition of language modeling to PyLaia would certainly improve performance. Based on this observation, we started working on adding language modeling to PyLaia.

\subsubsection{Building n-gram language models}
Two well-known libraries for the development of n-gram language models are SRILM \cite{SRILM} and KenLM \cite{kenlm}. 
The SRILM toolkit provides a comprehensive solution for language modelling, including n-gram models. SRILM provides a set of utilities for building, training and applying language models. Widely used in both academia and industry, SRILM's flexibility and robust performance make it ideal for a variety of language processing tasks, including ASR and machine translation.
KenLM is a widely used and efficient tool for building and using n-gram language models. It provides functionalities for training models on large text corpora and has fast and memory efficient decoding capabilities. KenLM's versatility extends to the support of different levels of tokenization, making it adaptable to various applications in Automatic Text Recognition (ATR) and Automatic Speech Recognition (ASR).
Both libraries are able to build language models at different levels: character, subword or word.

\subsubsection{Decoding with language models}
PyLaia's basic decoding is achieved with the best path decoding algorithm, which selects the token with the highest probability at each time step. 
In contrast, decoding with a language model is done with a beam search decoding algorithm. This method combines the probabilities from the optical model with the conditional probabilities of the language model, taking into account lexicon constraints. 
We rely on torchaudio's \texttt{ctc\_decoder}\footnote{\url{https://pytorch.org/audio/main/generated/torchaudio.models.decoder.ctc\_decoder.html}} function to combine PyLaia and n-gram models. This implementation supports language models built at any level (characters, subwords or words).

\subsection{Resources}

We have ensured that resources are available to users and contributors.

We provide a comprehensive user documentation for PyLaia\footnote{\url{https://atr.pages.teklia.com/pylaia/usage}}, which includes guidelines for formatting datasets, creating and training models, and predicting on unseen images. Users will find detailed explanations of each parameter, ensuring a clear understanding of its functionality. The documentation also provides numerous examples to illustrate the use of different commands and parameters.

We also provide a contributor documentation\footnote{\url{https://atr.pages.teklia.com/pylaia/get_started/development/}} for those interested in contributing to the codebase. This resource guides contributors on how to effectively file issues and suggest enhancements. It also provides instructions for developers who wish to contribute to code improvements, bug fixes or new features.

Finally, we release a hub of 12 pre-trained models\footnote{\url{https://huggingface.co/Teklia/}} on Hugging Face. Other users are also encouraged to publish open-source models.
To ensure reproducibility, all datasets used in this study are also shared in Hugging Face. Datasets are stored as Parquet\footnote{\url{https://parquet.apache.org}} files, which include the images and their associated ground truth transcription. Additionally, we provide a PyLaia command to easily convert them in PyLaia format. 

\section{Experimental setup}
\label{sec:experimental-setup}

We perform several experiments to measure how our contributions affect PyLaia. First, we compare different ways of computing confidence scores and measure the impact of calibration techniques. Second, we measure the improvement achieved by PyLaia when using language modeling. 
This section introduces the architecture, training configuration, and various datasets used in this work.

\subsection{Model}

The same architecture is used in all experiments. Four convolutional blocks are followed by three recurrent blocks. 
Each convolutional block uses a kernel size of 3, with a pooling size set to 2, except for the third block where it is set to 0. The number of features increases at each block: 12, 24, 48, 48. Batch normalization is applied and the LeakyReLU activation function is used. 
Recurrent blocks are a bidirectional LSTM layer with 256 units. Dropout is used with $p=0.5$. 
The configuration file for this architecture is included in the documentation. 

Text lines are resized to a height of 128 pixels and random affine transformations are applied during training. Models are trained with a batch size of 8. Early stopping is used with a patience of 80 epochs. A learning rate scheduler is also used to divide the learning rate by 10 after 5 epochs without improvement.

For each dataset, a 6-gram language model is built with modified Kneser-Ney smoothing \cite{smoothingLM} on the training set. We use a 6-gram model with a weight set to 1.5, as these were the optimal values measured on the NorHand v3 dataset.

\subsection{Datasets}

We present the 12 datasets used in our experiments. These datasets cover a large period of time, diverse writing styles and 7 languages, as presented in Table. \ref{tab:dataset_split}.

\setlength{\tabcolsep}{5pt}
\begin{table}[]
    \centering
    \caption{Description of the different datasets used in this study.}
    \begin{tabular}{l|llrrr}
    \toprule
    \multirow{2}{*}{\textbf{Dataset}} & \multirow{2}{*}{\textbf{Language}} & \multirow{2}{*}{\textbf{Period}} & \multicolumn{3}{c}{\textbf{Number of lines}} \\
     & & & \textbf{Train} & \textbf{Val} & \textbf{Test} \\
    \midrule
    Himanis \cite{himanis-database} & Latin, French & medieval & 18,504 & 2,367 & 2,241 \\
    HOME-Alcar \cite{home-alcar-database} & Latin & medieval & 59,969 & 7,905 & 6,932  \\
    NewsEye \cite{newseye-readocr-austrian-database} & Austrian & 19th century & 38,891 & 3,282 & - \\
    NorHand v1 \cite{HuMu-opensourceHTR} & Norwegian & 19th century & 19,653 & 2,286 & 1,793 \\
    NorHand v2 \cite{humu-v2} & Norwegian & 19th century & 145,061 & 14,980 & 1,793  \\
    NorHand v3 \cite{humu-v3-database} & Norwegian & 19th century & 223,971 & 22,811, & 1,573  \\
    Belfort \cite{belfort-database} & French & 19th century & 25,800 & 3,102 & 3,819 \\
    Esposalles \cite{esposalles-database}  & Catalan & 19th century & 2,328 & 742  & 757 \\
    POPP (generic) \cite{popp-database}  & French & 19th century & 3,835 & 480  & 479 \\
    RIMES \cite{RIMES-database}  & French & modern & 10,188 & 1,138  & 778\\
    IAM \cite{IAM-database} & English & modern & 6,482 & 976 & 2,915 \\ 
    CASIA \cite{casia-hwdb2-database} & Chinese & modern & 33,425 & 8,325 & 10,449 \\
    \bottomrule
    \end{tabular}
    \label{tab:dataset_split}
\end{table}
 
\subsubsection{Himanis} (Historical MAnuscript Indexing for user controlled Search) \cite{himanis-database} is a corpus of medieval documents written in Latin or in French. The dataset is publicly available in Arkindex\footnote{\url{https://demo.arkindex.org/browse/5000e248-a624-4df1-8679-1b34679817ef?top_level=true&folder=true}}.

\subsubsection{HOME-Alcar}
The HOME-Alcar (Aligned and Annotated Cartularies) \cite{home-alcar-database} dataset is a Medieval corpus written in Latin. The 17 medieval manuscripts in this corpus are cartularies, i.e. books copying charters and legal acts, produced between the 12th and 14th centuries. 

\subsubsection{Newseye} The dataset \cite{newseye-readocr-austrian-database} comprises Austrian newspaper pages from 19th and early 20th century. The images were provided by the Austrian National Library and include 148 training pages and 13 pages for validation. This is the only dataset in our study that contains exclusively printed text (Fraktur). Since there is no test set in this dataset, we provide results on the validation set.

\subsubsection{NorHand (v1, v2, v3)} The NorHand datasets \cite{HuMu-opensourceHTR,humu-v2,humu-v3-database} are composed of Norwegian handwritten letters and diaries from 19th and early 20th century. Three versions are available, with an increasing number of annotated pages. The test set of the second and third version is more diverse and difficult, as each test page was written by a different writer. 

\subsubsection{Belfort} The Belfort dataset \cite{belfort-database} includes minutes of the municipal council of the French city of Belfort. The handwritten documents are drawn up between 1790 and 1946, and include deliberations, lists of councilors, convocations, and agendas. We use the random split in our experiments.


\subsubsection{Esposalles} The ESPOSALLES dataset \cite{esposalles-database} is a collection of historical marriage records from the archives of the Cathedral of Barcelona. The corpus is composed of 125 pages. All documents are written in old Catalan by a single writer. It includes 125 pages with word, line and record segmentations.

\subsubsection{POPP} The POPP dataset \cite{popp-database} contains tabular documents from the 1926 Paris census. It contains 160 pages written in French, each page contains 30 lines. Each row is divided in 10 columns: surname, name, birthdate, birthplace, nationality, civil status, link, education level, occupation, and employer.

\subsubsection{RIMES} The RIMES dataset \cite{RIMES-database} is composed of administrative documents written in French. We use the \textit{Letters} subset that includes 3,700 training pages with corresponding transcriptions. Experiments are carried out on the RIMES-2011 version which includes only text lines from the letters' body (address, reference numbers, and signatures are excluded). 

\subsubsection{IAM} The IAM dataset \cite{IAM-database} is composed of modern documents in English, written by 500 writers. It includes 747 training pages with corresponding transcriptions. 
For our experiments, we use the RWTH partition (IAM-A or Aachen) of the dataset.

\subsubsection{CASIA-HWDB2} The CASIA-HWDB2 dataset \cite{casia-hwdb2-database} contains handwritten Chinese documents. The dataset includes 3,261 training pages, 815 validation pages, and 1,015 test pages.

\section{Results}
\label{sec:results}

In this section, we present and discuss the results of our experiments. 
First, we analyze different methods for reliable confidence score estimation. 
We then discuss how adding language models affects the performance and speed of PyLaia.

\subsection{Confidence score analysis}
First, we compare different confidence estimation methods by measuring the correlation between confidence scores and recognition rate. Then, we discuss the benefit of confidence calibration using temperature scaling. 

\subsubsection{Best method for confidence estimation}
\label{sec:res_conf}

Table \ref{tab:correlation} presents the correlation between different computation methods for confidence scores and the recognition rate, defined as $1-CER$. Both Spearman (rank correlation) and Pearson (linear correlation) coefficients are computed to capture different aspects of the correlation.

\begin{table}[tbh]
    \centering
    \caption{Correlation of different computation methods for confidence scores with the recognition rate. Experiments are carried out on the NorHand v1 dataset. In all experiments, the p-value is always well below 0.05, indicating that the correlations are always significant.}
    \label{tab:correlation}
    \begin{tabular}{l|cc}
        \toprule
        \textbf{Computation method} & \textbf{Rank correlation} & \textbf{Linear correlation} \\
        \midrule
        Softmax mean (before CTC) & 0.52 & 0.56\\
        Softmax mean (after CTC) & \textbf{0.60} & \textbf{0.62}\\
        Softmax difference &	0.49 &	0.53\\
        Softmax entropy  &	0.32 &	0.34 \\
        \bottomrule
    \end{tabular}
\end{table}

Unfortunately, we find that the Monte Carlo dropout approach leads to a degradation of recognition quality in addition to high computational cost. Therefore, this method is excluded from our analysis.

The results show that the most effective method is to take the softmax value for each token after the CTC decoding step and average these values for each line to obtain line-level confidence scores. This approach not only ensures computational efficiency, but also yields a strong correlation with the recognition rate. As a result of these findings, we have incorporated this calculation method into PyLaia.

\subsubsection{Confidence score calibration}
\label{sec:res_calibration}


The optimal temperature $T$ is determined using a grid search algorithm on the validation set (from 1 to 6 with a step of 0.5). Once $T$ is found, we compute the correlation between confidence scores and the recognition rate on the test set, with and without temperature scaling. 

Table \ref{tab:calibration_correlation} provides insight into the impact of model calibration on the linear correlation between confidence scores and recognition rate on the test set of different datasets.

\begin{table}[tbh]
    \centering
    \caption{Impact of model calibration using the optimal temperature \textit{T} on the correlation between confidence scores and the recognition rate. In all experiments, the p-value is always well below 0.05, indicating that the linear correlation is always significant.}
    \label{tab:calibration_correlation}
    \begin{tabular}{l|ccc}
    \toprule
    \multirow{2}{*}{\textbf{Dataset}} & \multirow{2}{*}{\textbf{$T$}} &  \textbf{Linear correlation} & \textbf{Linear correlation} \\
    &   &  \textbf{without scaling}  & \textbf{with scaling}  \\
    \midrule 
    Himanis & 2.0 & 0.64 & \textbf{0.70} \\
    HOME-Alcar & 2.0 & 0.55 &  \textbf{0.59} \\
    NewsEye* & 2.5 & 0.45 & \textbf{0.74}  \\
    NorHand v1 & 2.0 & 0.62 & \textbf{0.79} \\
    NorHand v2 & 2.0 & 0.43 & \textbf{0.69} \\
    NorHand v3 & 2.0 & 0.71 & \textbf{0.78} \\
    Belfort & 4.0 & 0.58 & \textbf{0.78} \\
    Esposalles & 3.5 & 0.23 & \textbf{0.71} \\
    POPP & 2.5 & 0.68 & \textbf{0.75} \\
    RIMES & 2.0 & 0.45 & \textbf{0.56} \\ 
    IAM & 2.5 & 0.63 & \textbf{0.72} \\ 
    CASIA & 1.5 & 0.68 & \textbf{0.77}  \\
    \bottomrule
    \multicolumn{4}{l}{* Evaluation done on the validation set.}
    \end{tabular}
    \label{tab:results}
\end{table}

The results show significant improvements in the linear correlation between confidence scores and recognition rates when temperature scaling is applied. Across different datasets, optimal temperatures ranging from 1.5 to 4.0 yield consistently higher correlations compared to scenarios without scaling. Since temperature scaling only scales the logits without changing the underlying ranking of the predictions, it does not affect the rank correlation.

Our analysis suggests that the default value of 2.0 is a robust choice for calibration, even for datasets where the optimal temperature is not equal to 2.0. For example, setting the temperature to 2.0 yields a correlation of 0.74 on the Belfort dataset (versus 0.78 for the optimal temperature of 4.0) and a correlation of 0.69 on the Esposalles dataset (versus 0.72 for the optimal temperature of 3.5).

\cite{pylaia}

\subsection{Impact of language modeling}

In this section, we investigate the influence of language models on PyLaia's performance for the task of automatic text recognition. 
We measure the Character Error Rate (CER) and the Word Error Rate (WER) with and without the language model, across 12 different datasets covering a wide range of languages and historical periods.

The results, presented in Table \ref{tab:results}, show that decoding with a language model leads to a significant performance improvement. On average, the CER is reduced by 11.9\% (0.9 absolute points), while the WER is reduced by 12.9\% (3.3 absolute points). This suggests that the linguistic constraints provided by the language model improve PyLaia's performance. 

\setlength{\tabcolsep}{5pt}
\begin{table}[]
    \centering
    \caption{Evaluation of PyLaia with and without a 6-gram character language model. Results are presented on 12 datasets (test set). $\Delta$ refers to the relative change (\%) between scores without language model and with language model.}
    \begin{tabular}{l|rrr|rrr|r}
    \toprule
    \textbf{Dataset} &  \multicolumn{3}{c}{\textbf{CER (\%)}} & \multicolumn{3}{|c|}{\textbf{WER (\%)}} & \multirow{2}{*}{\textbf{N lines}} \\
    \textbf{Language model}   & \textbf{No} & \textbf{Yes} & \textbf{$\Delta$} & \textbf{No} & \textbf{Yes} & \textbf{$\Delta$} & \\
    \midrule
    Himanis & 9.87 & \textbf{8.87} & \color{green}-10.13 & 29.25 & \textbf{24.37} & \color{green}-16.68 & 2,241 \\
    HOME-Alcar & 8.35  & \textbf{7.85} & \color{green}-5.99 & 26.15 & \textbf{23.20} & \color{green}-11.28 & 6,932 \\
    NewsEye* & 1.82 & \textbf{1.77} & \color{green}-2.75& 7.77 & \textbf{7.01} & \color{green}-9.78 & 4,667 \\
    NorHand v1 &  7.94 & \textbf{6.55} & \color{green}-17.51 & 24.04 & \textbf{18.20} & \color{green}-24.29 & 1,793  \\
    NorHand v2 &  9.72 & \textbf{8.23} & \color{green}-15.33& 27.78  & \textbf{21.65} & \color{green}-22.07 & 1,562  \\
    NorHand v3 &  7.52 & \textbf{6.36} & \color{green}-15.43& 22.99  & \textbf{18.11} & \color{green}-21.23 & 1,562  \\
    Belfort & 10.54&  \textbf{9.52} & \color{green}-9.68 & 28.12  & \textbf{23.73} & \color{green}-15.61 & 3,819 \\
    Esposalles  & \textbf{0.76} & 1.04 & \color{red}36.84 &  \textbf{2.62} & 3.38 & \color{red}29.01 & 757 \\
    POPP  & 16.49 & \textbf{16.09} & \color{green}-5.70 & 36.26 & \textbf{34.52} & \color{green}-8.23 &	479 \\
    RIMES  & 4.55 & \textbf{3.82} & \color{green}-16.04 & 14.39  & \textbf{10.53} & \color{green}-26.82 & 778 \\
    IAM  & 8.44 & \textbf{7.50} & \color{green}-11.14 & 24.51 & \textbf{20.98} & \color{green}-14.40 & 2,915 \\ 
    CASIA & 4.61 & \textbf{1.53} & \color{green}-66.81 & - & - & - & 10,449 \\
    \bottomrule
    \multicolumn{4}{l}{* Evaluation done on the validation set.}
    \end{tabular}
    \label{tab:results}
\end{table}

Despite the overall improvement, there are differences between the datasets, especially depending on the script and writing style. For example, the language model drastically reduces the CER on the CASIA dataset. This large gain is likely due to the logographic nature of Chinese writing, where each symbol represents either a word or a minimal unit of meaning. The gain (3 points) corresponds to the average WER gain on latin-script datasets.

On the other hand, for datasets with already very low error rates (CER $< 2\%$), the impact of the language model is either limited or negative, as for the NewsEye and Esposalles datasets respectively. The good performance of PyLaia without language model is probably due to the relative simplicity of these datasets, as the writing style is uniform (printed text for NewsEye, single writer for Esposalles). In this case, we observe that the language model tends to create more errors than it can correct.

Finally, for all the other datasets, the language model is beneficial, with an average reduction of 0.9 points in CER (12.2\% relative improvement) and 4.0 points in WER (17.9\% relative improvement). Overall, language modeling improves performance on 11 datasets out of the 12 that we tested. One of the key advantage of n-gram language models is that they can be built and used easily without any expert knowledge, and without requiring any additional data. This improvement in performance comes at the cost of increased computation time, as analyzed in the appendix \ref{sec:speed}.

\subsection{Comparison with state-of-the-art}

\setlength{\tabcolsep}{5pt}

In Table \ref{tab:benchmark} we present a benchmark comparing PyLaia with state-of-the-art models on different datasets. Four datasets with reference evaluation are selected: NorHand v1, IAM, Rimes, and POPP.

On NorHand v1, PyLaia shows competitive performance compared to other systems, showing a significant reduction in both CER and WER. In addition, on Rimes, PyLaia performs slightly below the transformer-based model, but achieves respectable performance. 
In contrast, on POPP, PyLaia falls well below the performance of the VAN model. We also observed degraded performance on other private datasets containing tables, suggesting that the CTC mechanism may not be well suited for such documents featuring blank areas and abbreviations.
Finally, on IAM, PyLaia falls below other models, especially those based on Transformer architectures. However, the existence of different partitions in IAM makes it difficult to perform a comprehensive analysis.

There are two major limitations of PyLaia.
First, its reliance on CTC, which limits training to line-level recognition rather than full document recognition. Second, PyLaia is not based on a transformer. As a result, it underperforms compared to more recent architectures such as DAN \cite{dan}, trOCR \cite{trOCR}, S-Attn \cite{rethinking-line-atr}, and VAN \cite{van}. 
Despite these limitations, PyLaia performs close to state-of-the-art on different datasets. Moreover, it is fast, efficient and does not require as much computing power as more recent architectures. Finally, PyLaia is easy to use, even for non-experts, and is already integrated in two document processing platforms: Transkribus and Arkindex. 
For these reasons, we believe that PyLaia is still relevant in the Document Analysis and Recognition (DAR) community.

\begin{table}[htb]
\begin{threeparttable}
    \centering
    \caption{Comparison of PyLaia with state-of-the-art models across multiple dataset. Best scores appear in bold.}
    \begin{tabular}{llllrr}
    \toprule
    \textbf{Dataset} & \textbf{Model} & \textbf{LM} & \textbf{Add. data} &  \textbf{CER (\%)} & \textbf{WER (\%)} \\
    \midrule
    NorHand v1 & Kraken expert \cite{HuMu-opensourceHTR} &   &  & 12.2         & 31.3 \\ 
               & HTR-Flor++ expert \cite{HuMu-opensourceHTR} &  &  & 11.0 & 29.7 \\
               & Kaldi expert \cite{HuMu-opensourceHTR} &   &  & 9.2          & 22.2 \\
               & PyLaia expert \cite{HuMu-opensourceHTR} &  &  & 8.9 & 23.8 \\
               & HTR+ expert \cite{HuMu-opensourceHTR} &  &  & 8.3 & 20.3 \\
               & PyLaia+LM (ours) & \cmark &  & \textbf{6.6} & \textbf{18.2} \\ \midrule
    POPP &  PyLaia+LM (ours) & \cmark  &  & 16.1 & 34.5 \\
        & VAN \cite{popp-database} &   &  & 7.1 & 19.1 \\
        & VAN student \cite{popp-database} &  & \cmark & \textbf{4.5} & \textbf{13.6} \\
        
    \midrule           
    RIMES & PyLaia \cite{laia} & \cmark &  & 4.4 & 12.2  \\
          & PyLaia+LM (ours) & \cmark  &  & 3.2 & 10.1 \\
            & MDLSTM \cite{Voigtlaender} & \cmark &  & 2.8 & 9.6 \\
            & SFR \cite{wigington2018} & \cmark &  & 2.1 & \textbf{9.3} \\
            & S-Attn / CTC  \cite{rethinking-line-atr} & \cmark & \cmark & \textbf{2.0} & - \\
            
    \midrule
    IAM     & PyLaia+LM (ours) \tnote{a} & \cmark  &  & 7.5  & 21.0 \\ 
            & VAN \cite{van} \tnote{b}&  &  & 4.5 & 14.6 \\
            & DAN \cite{key-value} \tnote{a} &   &  & 4.3 & 13.7 \\
            & MDLSTM \cite{Voigtlaender} \tnote{b} & \cmark &  & 3.5 & \textbf{9.3} \\
            & TrOCR large \cite{trOCR} \tnote{b} &  & \cmark & 2.9 & - \\
            & S-Attn / CTC \cite{rethinking-line-atr} \tnote{b} & \cmark & \cmark & \textbf{2.8} & - \\
    \midrule
    \end{tabular}
    \label{tab:benchmark}
\begin{tablenotes}
\item[a] IAM-A/RWTH/Aachen; \item[b] IAM-B/original \\
\end{tablenotes}
\end{threeparttable}
\end{table}

\section{Conclusion}
\label{sec:conclusion}

In this article, we present our contributions to the PyLaia open-source automatic text recognition (ATR) library. First, we compare different confidence estimation methods and show that temperature scaling helps to obtain reliable confidence scores, thus improving the interpretability of predictions. Second, we propose a decoding mechanism that incorporates language models into the PyLaia library, further extending its capabilities. Finally, to ensure that other users can take advantage of these new features, we have written an extensive documentation that describes in details how to use or contribute to the library. We also release 12 trained models on the Hugging Face Hub.
Our work highlights the importance of maintaining ATR libraries for sustained progress in the field. As noted by Maarand \textit{et al.} \cite{HuMu-opensourceHTR}, many open-source ATR libraries are at risk of becoming obsolete due to lack of maintenance and contributors.

In the future, we plan to integrate PyLaia into the Hugging Face ecosystem. Our first initiative is to ensure users can seamlessly pull or publish datasets and models from or to the hub.
In addition, we also plan to establish a benchmark for ATR within a dedicated Hugging Face space. This initiative will enable automatic evaluation of any model available on the Hub across a comprehensive set of datasets. By creating this benchmark, we aim to provide a standardized and efficient means for researchers to evaluate the performance of various ATR models.



\section*{Acknowledgement}
We thank Joan Puigcerver and Carlos Mocholí for implementing PyLaia and for allowing us to contribute. We also thank Stefan Weil for his recent contributions to PyLaia. 
This work was supported by the Research Council of Norway through the 328598 IKTPLUSS HuginMunin project. 


\bibliographystyle{splncs04}
\bibliography{main}

\section*{Apprendix}

\subsection{Impact of language models on speed}
\label{sec:speed}

During the inference process, computations can take advantage of GPU acceleration up to the language model decoder, which is unfortunately limited to CPU execution. As a result, the inclusion of a language model in the decoding process improves the overall results, but results in a significant slowdown of the prediction speed. Specifically, the decoding speed of PyLaia experiences a tenfold reduction when a language model is applied, as detailed in the Table \ref{tab:speed}. Therefore, we recommend using language models for batch processing of documents, while cautioning against their use in real-time scenarios due to the associated computational overhead.

\begin{table}[]
    \centering
    \caption{Impact of language models on decoding time (s/image). Experiments are carried out on an NVIDIA GeForce RTX 3080 Ti GPU.}
    \label{tab:speed}
    \begin{tabular}{llc}
    \toprule
    \multirow{2}{*}{Device} & \multirow{2}{*}{Language model} & Speed  \\
     &  & (seconds/image)  \\
    \midrule
    GPU & No  & \textbf{0.01} \\
    CPU & No & 0.03 \\
    CPU & Yes & 0.12 \\
    \bottomrule
    \end{tabular}
\end{table}

\end{document}